%% file: main.tex
\documentclass[conference]{IEEEtran}
\IEEEoverridecommandlockouts
\usepackage{cite}
\usepackage{amsmath,amssymb,amsfonts}
\usepackage{algorithmic}
\usepackage{graphicx}
\usepackage{textcomp}
\usepackage{xcolor}
\usepackage{url}
\usepackage{pifont}
\usepackage{microtype}
\usepackage{hyperref}

\def\BibTeX{{\rm B\kern-.05em{\sc i\kern-.025em b}\kern-.08em
    T\kern-.1667em\lower.7ex\hbox{E}\kern-.125emX}}
\begin{document}

\title{Advancing Behavior Generation in Mobile Robotics through High-Fidelity Procedural Simulations}

\author{\IEEEauthorblockN{Victor A. Kich*}
\IEEEauthorblockA{\textit{Intelligent Robot Laboratory}\\
\textit{University of Tsukuba}\\
Tsukuba, Japan\\
Email: victorkich98@gmail.com}
\and
\IEEEauthorblockN{Jair A. Bottega*}
\IEEEauthorblockA{\textit{Intelligent Robot Laboratory}\\
\textit{University of Tsukuba}\\
Tsukuba, Japan\\
Email: jairaugustobottega@gmail.com}
\and
\IEEEauthorblockN{Raul Steinmetz}
\IEEEauthorblockA{\textit{Centro de Tecnologia}\\
\textit{Universidade Federal de Santa Maria}\\
Santa Maria, Brazil\\
Email: rsteinmetz@inf.ufsm.br}
\and
\IEEEauthorblockN{Ricardo B. Grando}
\IEEEauthorblockA{\textit{Robotics and AI Lab}\\
\textit{Technological University of Uruguay}\\
Rivera, Uruguay\\
Email: ricardo.bedin@utec.edu.uy}
\and
\IEEEauthorblockN{Ayanori Yorozu}
\IEEEauthorblockA{\textit{Intelligent Robot Laboratory}\\
\textit{University of Tsukuba}\\
Tsukuba, Japan\\
Email: yorozu@cs.tsukuba.ac.jp}
\and
\IEEEauthorblockN{Akihisa Ohya}
\IEEEauthorblockA{\textit{Intelligent Robot Laboratory}\\
\textit{University of Tsukuba}\\
Tsukuba, Japan\\
Email: ohya@cs.tsukuba.ac.jp}
\thanks{*~Victor A. Kich and Jair A. Bottega contributed equally.}
}

\maketitle

\input{sessions/0_abstract}
\input{sessions/1_introduction}
\input{sessions/2_related_works}
\input{sessions/3_methodology}
\input{sessions/4_results}
\input{sessions/5_conclusions}

\bibliographystyle{IEEEtran}
\bibliography{main}

\end{document}

%% file: sessions/0_abstract.tex
\begin{abstract}
This paper introduces YamaS, a simulator integrating Unity3D Engine with Robotic Operating System for robot navigation research and aims to facilitate the development of both Deep Reinforcement Learning (Deep-RL) and Natural Language Processing (NLP). It supports single and multi-agent configurations with features like procedural environment generation, RGB vision, and dynamic obstacle navigation. Unique to YamaS is its ability to construct single and multi-agent environments, as well as generating agent's behaviour through textual descriptions. The simulator's fidelity is underscored by comparisons with the real-world Yamabiko Beego robot, demonstrating high accuracy in sensor simulations and spatial reasoning. Moreover, YamaS integrates Virtual Reality (VR) to augment Human-Robot Interaction (HRI) studies, providing an immersive platform for developers and researchers. This fusion establishes YamaS as a versatile and valuable tool for the development and testing of autonomous systems, contributing to the fields of robot simulation and AI-driven training methodologies.
\end{abstract}

\begin{IEEEkeywords}
Robot Navigation, Simulator, Deep Reinforcement Learning, Natural Language Processing, Procedural Generation, Virtual Reality
\end{IEEEkeywords}

%% file: sessions/1_introduction.tex
\section*{Suplementary Material}

The framework developed in this research is available for public access at the following link: \url{https://github.com/victorkich/YamaS}.

\section{Introduction}\label{section:introduction}

Robotics has evolved significantly over the past few decades, transitioning from simple, manually programmed machines to complex autonomous systems capable of learning and adapting to their environment. This evolution has been propelled by advancements in computing power, sensor technology, and artificial intelligence, particularly in the domain of deep learning~\cite{Goodfellow-et-al-2016}. However, developing and testing these intelligent systems in the real world poses substantial challenges, including safety risks and the high cost of iterative experimentation. Simulation environments provide a compelling alternative, offering a safe, cost-effective, and highly flexible platform for developing, testing, and refining robotic behaviors~\cite{Brooks1986, Koenig2004}.

Simulators in robotics serve multiple purposes, from validating theoretical models to training Artificial Intelligent (AI) agents via Reinforcement Learning (RL). The integration of realistic physics engines and the ability to simulate complex sensors, such as LiDAR and RGB cameras, have made simulators an indispensable tool in robotics research~\cite{Koenig2004, Todorov2012}. Furthermore, the advent of simulators capable of handling multi-agent systems has opened new avenues for research in cooperative robotics and swarm intelligence~\cite{Stone2000, Dorigo2014}.

\begin{figure}[tp!]
    \centering
    \includegraphics[width=\linewidth]{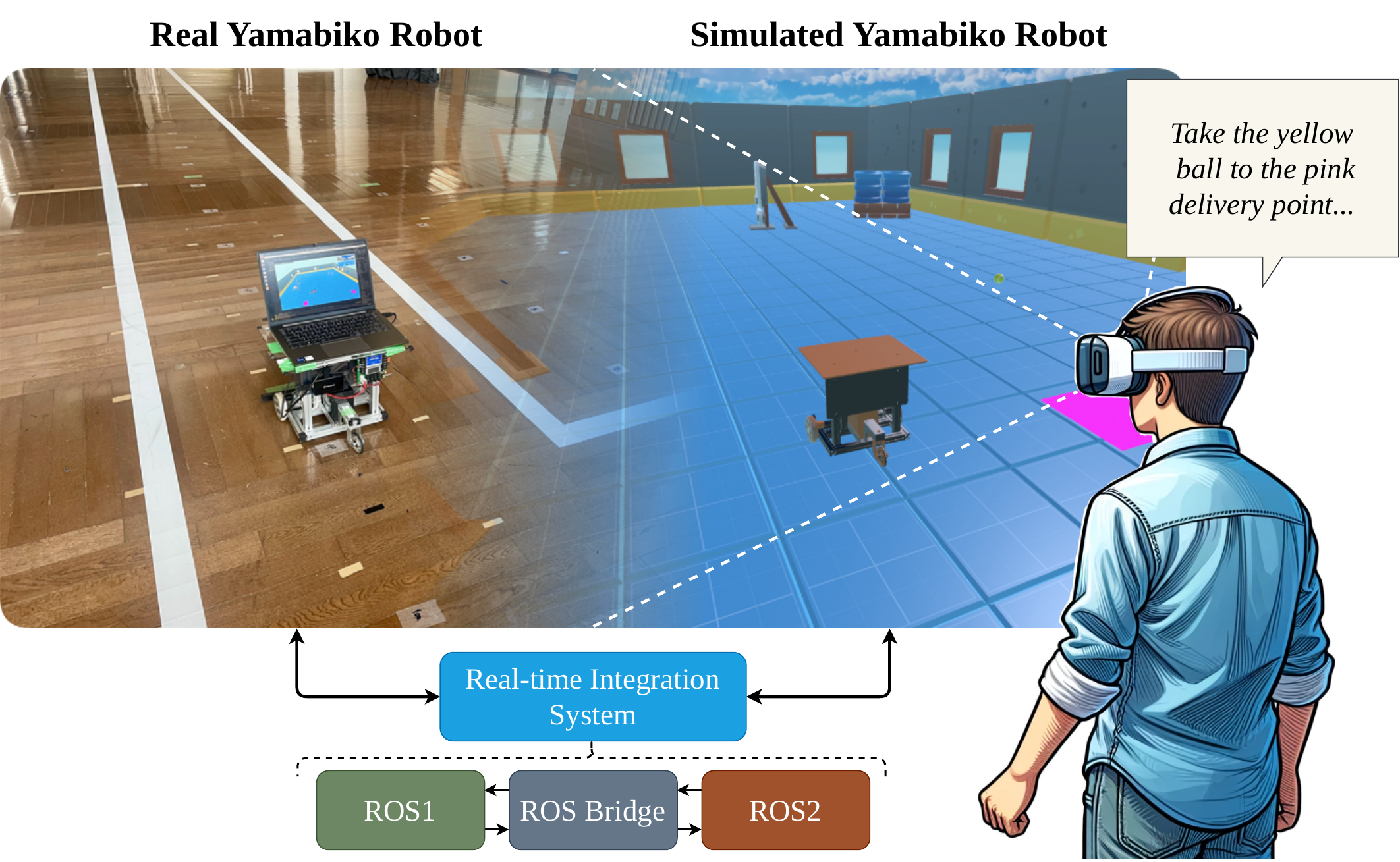}
    \caption{Interplay between Reality and Simulation: The Yamabiko Beego robot in action alongside its simulated avatar, with a researcher in VR guiding the task for behavior generation. This visual encapsulates the seamless real-time integration achieved through ROS1 and ROS2, pivotal for synchronizing the physical and virtual realms to refine autonomous navigation strategies.}
    \label{fig:project_diagram}
    \vspace{-5mm}
\end{figure}

Despite these advancements, a gap remains between simulated environments and the unpredictable nature of the real world, often referred to as the ``reality gap"~\cite{Jakobi1995}. Another common problem is that creating diverse scenarios and programming experiments remains a challenging and laborious task for most simulators, highlighting a continued need for tools that streamline and accelerate the development of environments conducive to robot learning~\cite{kaur2022simulators}.

This work introduces YamaS, a novel simulator designed to bridge the gap between virtual training and real-world application for autonomous robotic navigation. Built on Unity3D and integrated with both Robot Operating System (ROS)~\cite{ros} and ROS2~\cite{ros2}, YamaS offers a versatile platform for both single and multi-agent scenarios, enhanced with features like procedural generation, Natural Language Processing (NLP) for environment design and Virtual Reality (VR). By leveraging Deep Reinforcement Learning (Deep-RL) and advanced simulation capabilities, YamaS aims to facilitate the development of robots that are not only adept at navigating complex environments but can also perform tasks collaboratively, responding dynamically to the challenges posed by their surroundings.

Our contributions to the field of robotics and simulation are manifold:
\begin{itemize}
    \item We present a high-fidelity simulation framework that supports both single and multi-agent environments, enabling detailed scenario configurations for comprehensive testing and development.
    \item YamaS integrates procedural generation and NLP to allow for the dynamic creation of complex, varied environments, fostering innovative approaches to AI training and robot behavior optimization.
    \item Through test validation with the Yamabiko Beego robot, we demonstrate the simulator's accuracy in replicating real-world sensor data and robot kinematics, significantly reducing the reality gap.
    \item The simulator is integrated with VR, aiming for more precise robot behaviour observation in the eyes of the researcher. 
    \item Our work paves the way for future research in autonomous systems, offering a robust platform for exploring collaborative strategies, Deep-RL algorithms, and the application of Large Language Models (LLMs) in robotic navigation.
\end{itemize}

%% file: sessions/2_related_works.tex
\section{Related Work}\label{section:related_works}


In the evolving landscape of robotics research, the synergy between simulations and advanced technologies plays a pivotal role in pushing the boundaries of what's achievable, both in terms of innovation and practical application. This section aims to weave a more cohesive narrative that underscores the critical interplay of simulations with procedural generation, LLMs and VR.

The Gazebo simulator \cite{koenig2004design} was one of the first physics environments introduced in the fields of Robotics. Throughout the years it had many improvements and plugins which made the simulator widely used for research in robotics and artificial intelligence. More recently the simulator was extensively used for research covering RL and LLMs applications in robotics.

Gazebo and other simulators show that physics simulators are indispensable tools in robotics research. Due to the prohibitive costs and practical limitations associated with physical robots, simulators offer a viable alternative for testing and validating theoretical methods. The diversity of simulators, each with its unique offerings, forms a rich ecosystem that supports a wide range of research activities~\cite{collins2021review}. Building upon this foundation, the work by Katara \textit{et al.}~\cite{katara2023gen2sim} introduces Generation to Simulation (Gen2Sim), a method that significantly enhances robot skill learning in simulations. By automating the generation of 3D assets using 2D maps and associated task descriptions using generative models such as LLMs, Gen2Sim opens new avenues for developing and testing complex tasks in a simulated environment. This innovation not only broadens the scope of tasks that can be simulated but also introduces a higher level of realism and diversity into the simulation process.

The examination of simulation environments is further enriched by Körber \textit{et al.}~\cite{korber2021comparing}, who compare the performance of various simulation platforms. Their work highlights the importance of choosing the right simulation environment and hardware configuration to optimize the training efficiency of RL agents. This aspect of simulation research underscores the technical considerations that must be taken into account to maximize the utility of simulations in robotics research.

The integration of LLMs into robotics, as discussed by Zeng \textit{et al.}~\cite{zeng2023large}, represents a significant leap forward. The ability of LLMs to process and generate language in a contextually relevant manner has vast implications for robotics, enabling more intuitive Human-Robot Interaction (HRI) and more sophisticated decision-making processes in robots. Expanding on the connection between LLMs and robots, Mai \textit{et al.}~\cite{mai2023llm} developed the LLM-Brain framework, utilizing LLMs as a robotic brain to integrate egocentric memory and control for robotic tasks. It employs a zero-shot learning approach and leverages multimodal language models, aiming to enhance robotic functionalities through improved memory and control mechanisms. Vemprala \textit{et al.}~\cite{vemprala2023chatgpt} investigated OpenAI's ChatGPT for robotics, combining design principles for prompt engineering and the creation of a high-level function library which allows ChatGPT to adapt to different robotics tasks, achieving impressive results from tasks ranging from basic reasoning to complex navigation and manipulation, facilitated by natural language interaction.

On the multi-agent area Kannan \textit{et al.}~\cite{kannan2023smart} introduced an innovative framework for multi-robot task planning using LLMs. It demonstrates how LLMs can be used for task decomposition, coalition formation, and task allocation in multi-robot systems, addressing the complexity of coordinating tasks among heterogeneous robots. The paper presents a benchmark dataset for evaluating multi-robot planning systems and shows promising results in both simulated and real-world scenarios.

Procedural modeling is a technique for semi-automatically generating content using predefined algorithms and processes. Its efficiency in data compression and capability to produce varied and intricate content with limited human intervention makes it highly advantageous. This method has gained traction in the creation of virtual environments for movies, video games, and simulation applications due to its significant benefits~\cite{smelik2014survey}. Utilizing these techniques, Medina \textit{et al.}~\cite{gonzalez2016procedural} highlighted the importance of procedural generation for environments in robotic simulations. They introduced a framework capable of generating procedural cities, demonstrating its usefulness and feasibility in simulations. In the same vein, Arnold and Alexander~\cite{arnold2013testing} applied procedural generation to develop a broad range of test scenarios for robotic applications. These scenarios were employed to observe and evaluate robot behavior under various conditions. Furthermore, they devised a system for assessing these scenarios based on their potential safety implications through an event-based scoring mechanism. Scenarios receiving high scores, which suggest possible hazardous behaviors, were identified for more detailed review by safety engineers. Such an approach considerably diminishes the necessity for manual creation and monitoring of test situations, streamlining the testing process.

Virtual Reality (VR) constitutes a computer-simulated environment that enables user interaction in a manner that mimics real-world experiences, facilitated by specialized equipment like headsets with integrated screens or sensor-equipped gloves. This immersive technology has found applications across a diverse range of fields, as documented in several key studies~\cite{kaminska2019virtual, xie2021review, moline1997virtual, dascal2017virtual}. Exploring the dynamic advancements in VR technology, Burdea~\cite{burdea1999invited} revealed the potential synergies between VR and robotics, underscoring their collective benefits across various domains. In an innovative application of VR, Bottega \textit{et al.}~\cite{bottega2022jubileo, bottega2023jubileo} developed a simulation framework to enhance the development and testing of humanoid robots, focusing on HRI. Furthermore, leveraging advanced VR and graphics rendering technologies, Gonzalez \textit{et al.}~\cite{martinez2020unrealrox} introduced a virtual environment designed to bridge the gap between virtual simulations and real-world scenarios. This environment features hyper-realistic indoor settings for exploration by both robotic agents and human researchers, aiming to facilitate the development of innovative solutions. These contributions underscore the significance and advantages of employing virtual reality in robotic simulations, similar to the approach taken in the present work.

Our work builds upon these foundational insights by introducing a simulator that seamlessly integrates procedural generation, LLMs, and VR. This integration represents a holistic approach to robotic simulation, offering unprecedented ease of use, development capabilities, and sensor-based realism for both single and multi-agent scenarios on mobile robotics. Our contributions aim to offer a versatile platform for the next generation of robotics research and development.

%% file: sessions/3_methodology.tex
\section{Methodology}\label{section:methodology}

This section presents the integration and simulation of the Yamabiko Beego robot within both virtual and real-world environments, employing real-time data exchange and modular deployment techniques. Our approach encompasses the development of single and multi-agent simulation environments through YamaS, utilizing procedural generation for dynamic scenario creation. Additionally, we incorporate VR to enhance interaction realism, particularly in HRI studies, aiming to offer a practical and scalable solution for robotics research.

\subsection{Yamabiko Robot}

The Yamabiko Beego stands as a well-developed robotic platform in the field of research, specifically engineered for navigation and operation within natural environments. It embodies the principles of biomimicry, boasting advanced navigation sensors, environmental recognition capabilities, and obstacle avoidance systems \cite{giralt1984mobile}. Equipped with a distinctive locomotion system, the Yamabiko can navigate a variety of terrains, from steep inclines to rugged surfaces, making it an ideal candidate for exploration and environmental monitoring tasks. In the simulation, the Yamabiko is equipped with a LIDAR sensor and a frontal camera, complementing its sophisticated locomotion system.

\begin{figure}[tp!]
    \centering
    \includegraphics[width=\linewidth]{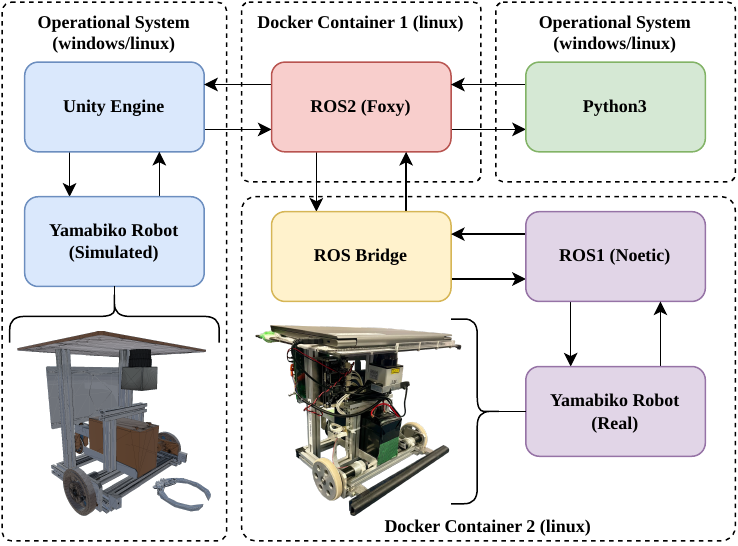}
    \caption{Architecture diagram showcasing the integration of the Unity Engine with ROS for the Yamabiko Beego robot simulation. This setup enables precise management and control of robotic behaviors and environmental dynamics through Python3 scripts, illustrating the seamless data exchange facilitated by the ROS Bridge between virtual and real-world components.}
    \label{fig:ros2}
    \vspace{-5mm}
\end{figure}

The integration and operation of the Yamabiko robot within both simulated environments and the real world, in real-time synchronization via ROS topics, are illustrated in Fig.~\ref{fig:ros2}. It is crucial to highlight that the agent can exchange data across both ROS1 and ROS2 frameworks, thanks to the utilization of the \emph{ros\_bridge}, facilitating data sharing over the network. Another notable aspect of our methodology is the project's modularization, using Docker containers to simplify deployment and utilization of the framework.

\subsection{Single-agent Environment}

\begin{figure*}[htbp!]
    \centering
    \includegraphics[width=\linewidth]{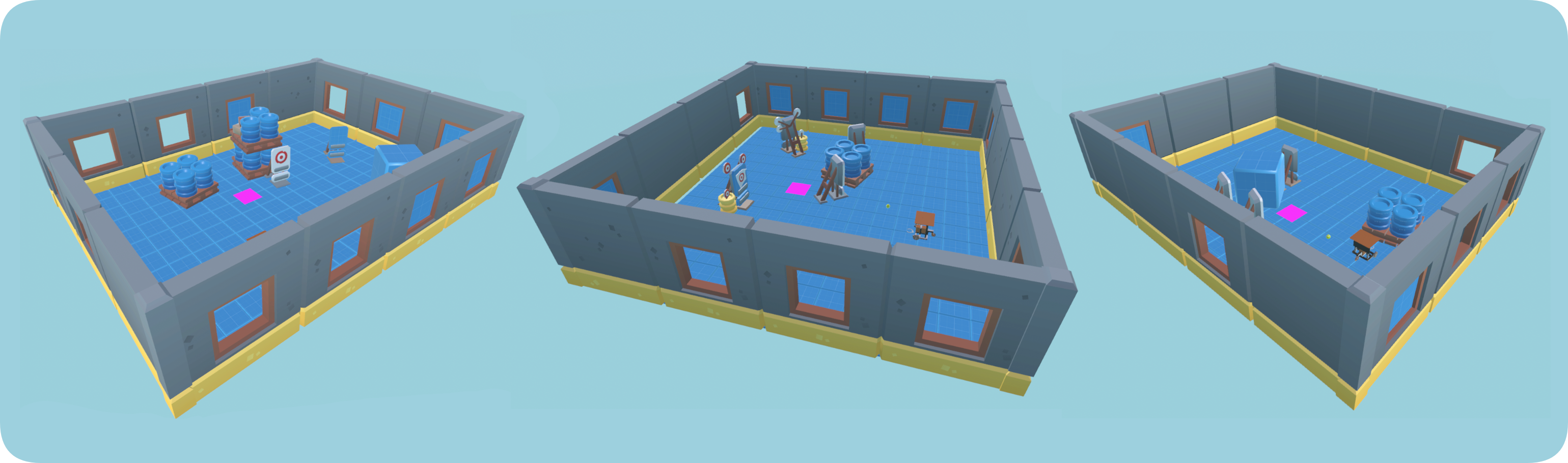}
    \caption{Example of a procedurally generated environments within the YamaS simulator. This visualization highlights the versatility of our simulation framework, where altering parameters via ROS topics dynamically alters the scene, introducing new challenges and scenarios for agent navigation and task execution.}
    \label{fig:procedural}
\end{figure*}

Within the YamaS simulator, the single-agent environment is intricately designed to support a broad spectrum of navigation and interaction tasks. This environment is highly customizable, allowing researchers to tailor the complexity and nature of tasks to their specific research goals. From elementary navigation puzzles to intricate interaction scenarios with dynamic obstacles and variable terrain, the environment offers unmatched flexibility. The procedural generation feature ensures each session presents a unique challenge, promoting robustness and adaptability in the tested algorithms. Examples of procedurally generated environments can be seen in Fig.~\ref{fig:procedural}.

Furthermore, a user interface menu enables the selection between single-agent and multi-agent scenarios, with parameters such as environment dimensions, obstacle count, ball count, and delivery zones being customizable. This adaptability ensures a tailored and dynamic simulation experience, conducive to a wide array of research objectives.

\subsection{Multi-agent Environment}

The multi-agent environment in YamaS focused to enhancing collaboration among agents, creating a complex landscape for the exploration of multi-agent system dynamics. This environment's complexity, with multiple interacting agents each pursuing their own objectives, ensures a rich variety of scenarios. Agents must adapt their strategies in real-time to succeed, making this setting ideal for examining emergent behaviors and cooperative task accomplishments.

\begin{figure}[htbp!]
    \vspace{-2mm}
    \centering
    \includegraphics[width=\linewidth]{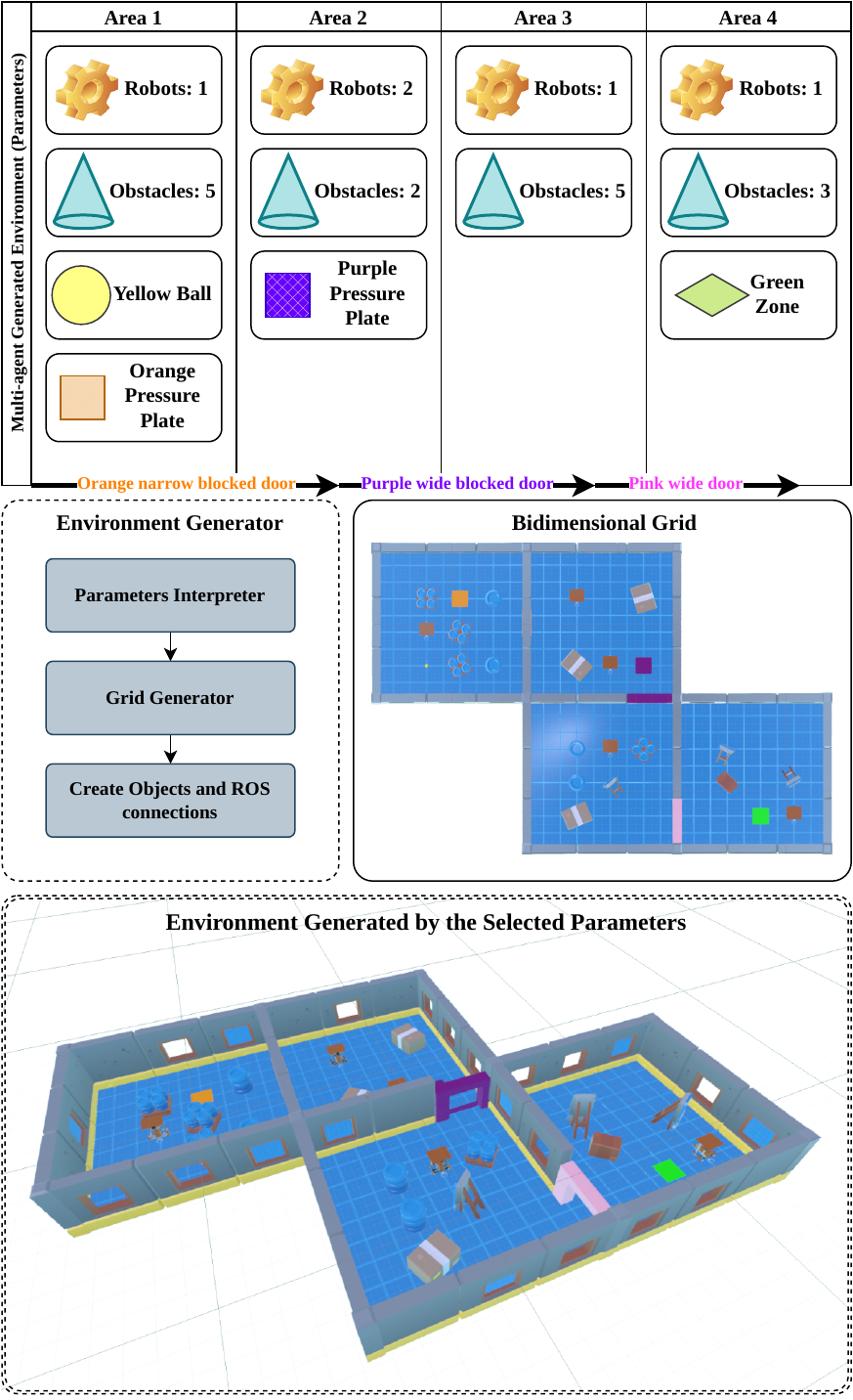}
    \caption{Flow diagram and visualization of YamaS's procedural environment setup for multi-agent simulations, showcasing parameter selection for areas, robots, and obstacles, leading to an interactive grid generation. The process culminates in a fully-formed simulation space, demonstrating the environment's readiness for multi-agent collaboration and task execution.}
    \label{fig:multi_agent}
\end{figure}

For configuring the multi-agent environment, the framework provides two methods: ROS topics and a user interface menu, allowing for detailed parameter adjustments such as the number of agents, specific objects, entry and exit definitions, color specifications, etc. Fig.~\ref{fig:multi_agent} illustrates a detailed example of a parameterized environment, structured and tested within a two-dimensional grid before the complete environment generation, including ROS connections and internal object spawning.

\subsection{Procedural Environment Generation}

YamaS harnesses procedural generation to create dynamic, versatile environments suited for single and multiple agent scenarios. This automated system select randomly between a preset of a list of objects, delivery areas, and balls, diversifying task environments with a wide range of sizes and orientations for simulation objects. Not only are the objects' positions procedurally generated, but the objects themselves also vary procedurally, enriching the simulation's complexity and realism.

The system organizes the environment into cells accommodating agents or objects, strategically placing interactive elements to ensure a dynamic and engaging simulation landscape. This dynamic update capability allows for real-time adaptations, ensuring simulations remain relevant and challenging.

Beyond mere object placement, YamaS actively generates a comprehensive description of the environment and disseminates this information via ROS topics, offering detailed insights into the terrain and the agents within it. Such descriptions allow for the application of LLMs to devise strategic behaviors for agents to execute tasks more effectively within the environment. This opens up possibilities for leveraging YamaS in various applications, including Deep-RL, LLM systems, and curriculum learning~\cite{bengio2009curriculum}.

\subsection{Virtual Reality Integration}

The VR integration significantly enhances the simulator, particularly in applications involving HRI. It provides researchers with an immersive platform for issuing voice commands and receiving responses from the robot, offering a more realistic testing environment for HRI applications. The VR interface not only facilitates a deeper understanding of the robot's actions and decisions by allowing real-time observation and interaction but also enhances the human's ability to monitor and adjust the robot's course as necessary. This immersive experience is crucial for optimizing the robot's navigation capabilities and aligning its performance with expected outcomes, offering a significant advancement in the development and testing of robotic systems.

%% file: sessions/4_results.tex
\section{Results}\label{section:experimental_results}

In this section, we present results gathered to validate the YamaS simulator's not only in the effectiveness in replicating real-world robotics behavior, but also the interaction and agent's behaviour generation through LLMs systems. Our experiments are designed to bridge the ``reality gap" by comparing the performance of simulated robotic tasks with their real-world counterparts. Through testing, we analyse the fidelity of the simulation and assess the extent to which virtual training can translate to tangible, real-world robotic skills.

\subsection{Sensor-based Reality Gap}

For make a comparison between the simulated and the real robot, it is necessary to create a movement function to generate the linear and angular velocities. This way, it is possible to generate a ground truth path based on the generated data.

To accomplish this task we proposed build a robot's movement function with an oscillating angular velocity and a constant linear speed. The mathematical expressions for the movement, particularly the angular velocity and the robot's position, are as follows:

\subsubsection{Angular Velocity}
The angular velocity $\omega(t)$ is modeled as a damped sinusoidal function that varies with time $t$:
\begin{equation}
\omega(t) = A \cdot e^{-\lambda t} \cdot \sin\left(\frac{2\pi (t - t_0)}{T}\right) + \text{bias},
\end{equation}
\noindent where:
\begin{itemize}
    \item $A$ is the amplitude of the sinusoidal wave, representing the maximum angular velocity.
    \item $\lambda$ is the damping factor that reduces the amplitude over time, providing smoother motion.
    \item $t_0$ is the initial movement time before the oscillation starts.
    \item $T$ is the time for a complete oscillation.
    \item The bias term compensates for any systematic deviation in the robot's movement.
\end{itemize}

The robot's position is updated based on its linear and angular velocities. The linear position along the $x$-axis, $x(t)$, and the $y$-axis, $y(t)$, are updated as follows:

\subsubsection{Linear Movement}
During the initial linear movement phase, the robot's desired position in the $x$-direction is incremented by:
\begin{equation}
\Delta x = v \cdot \Delta t,
\end{equation}
\noindent where $v$ is the constant linear speed and $\Delta t$ is the time step.

\subsubsection{Oscillatory Movement}
After the initial movement, both $x(t)$ and $y(t)$ are updated based on the robot's orientation, which is influenced by its angular velocity:
\begin{align}
x(t) &= x(t - \Delta t) + v \cdot \cos(\theta(t)) \cdot \Delta t, \\
y(t) &= y(t - \Delta t) + v \cdot \sin(\theta(t)) \cdot \Delta t,
\end{align}
\noindent where $\theta(t)$ represents the robot's orientation, adjusted by the angular velocity and normalized to ensure it remains within the range $[-\pi, \pi]$.

\subsubsection{Orientation Normalization}
The robot's orientation $\theta(t)$ is kept within the $[-\pi, \pi]$ range using the normalization:
\begin{equation}
\theta(t) = ((\theta(t) + \pi) \mod 2\pi) - \pi.
\end{equation}

This mathematical formulation enables the robot to execute a controlled oscillatory motion, combining linear advancement with angular adjustments to navigate through its environment.

\begin{figure}[tp!]
    \centering
    \includegraphics[width=\linewidth]{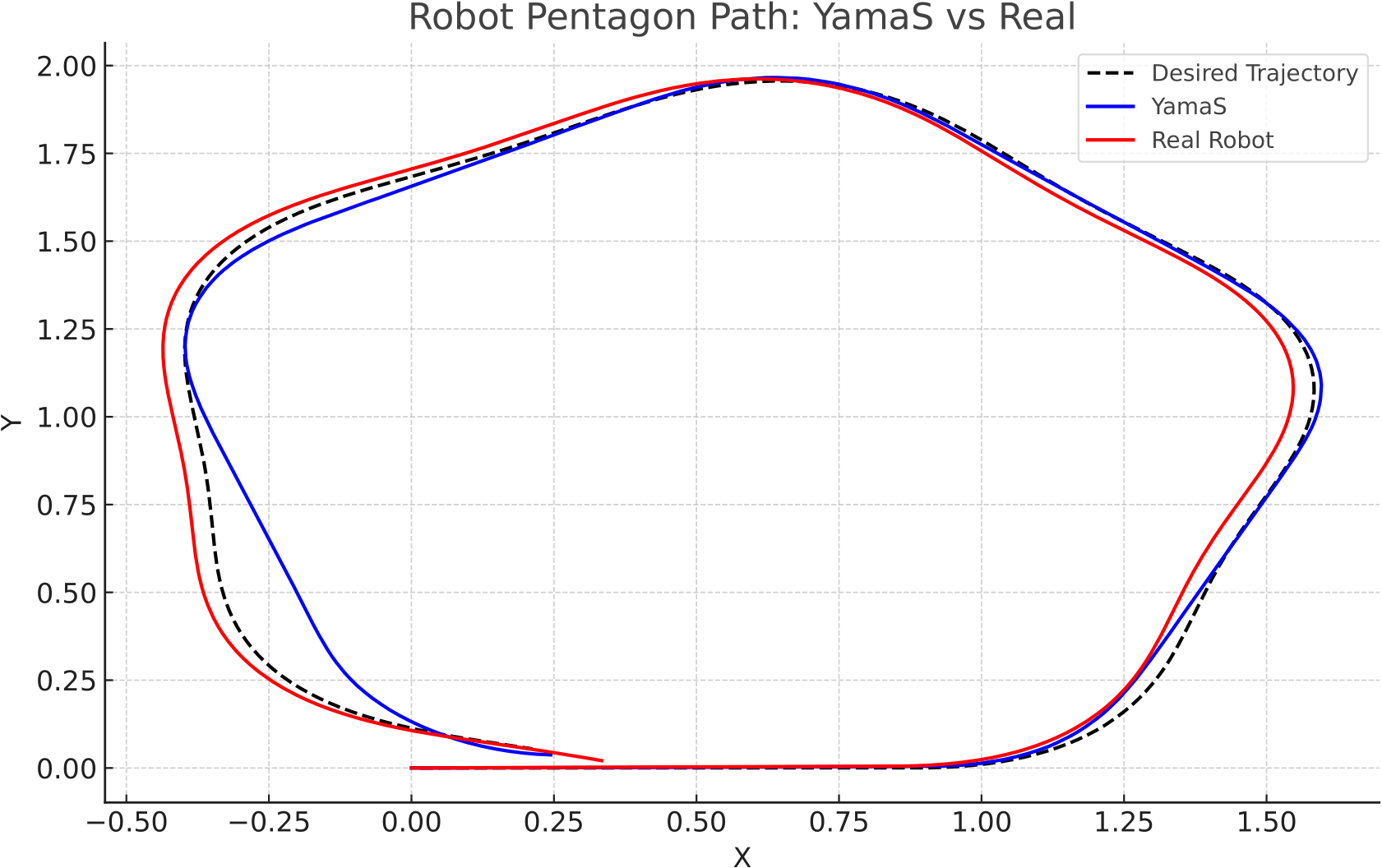}
    \caption{Comparative real-time odometry of the Yamabiko robot showcasing the pentagon path followed by the real robot (in red) against the desired trajectory (dashed line) and the simulated path (in blue), demonstrating the YamaS simulator's precision in mimicking actual robotic movement.}
    \label{fig:odometry_chart}
    \vspace{-5mm}
\end{figure}

\begin{figure*}[tp!]
    \centering
    \includegraphics[width=\linewidth]{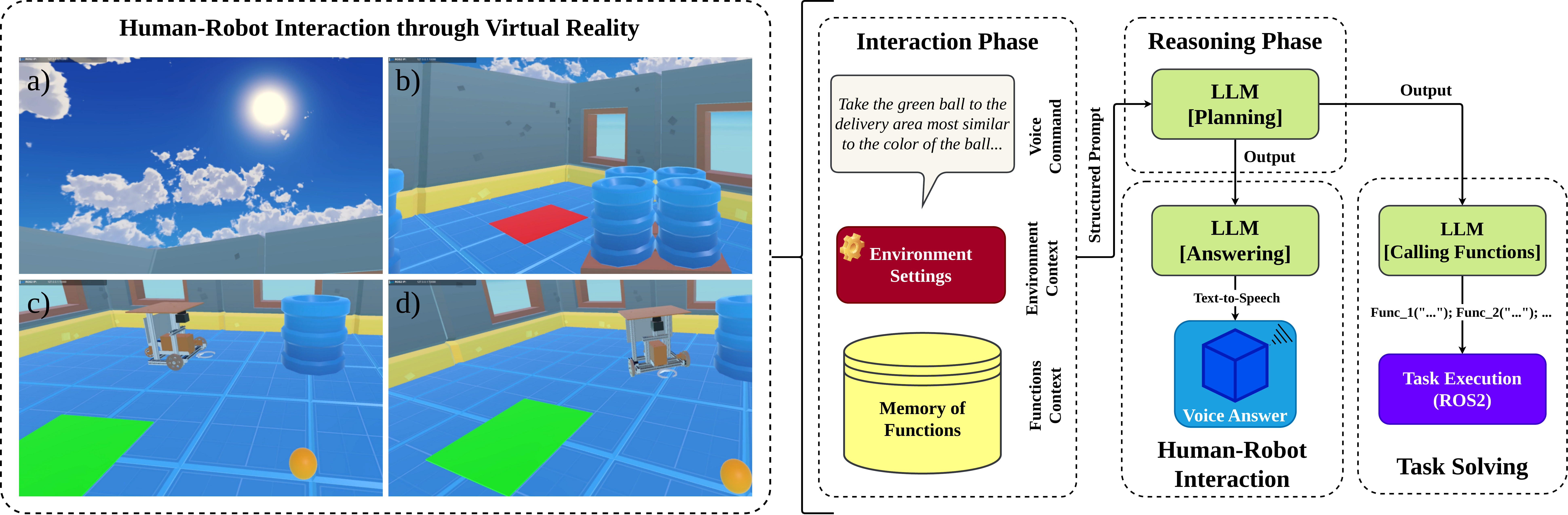}
    \caption{Sequence of human-robot interaction in VR, detailing the command and response flow: from initial voice prompt to the LLM agent's reasoning and subsequent task execution within the YamaS environment, illustrating the integration of verbal commands into actionable robotic behavior.}
    \label{fig:virtual_reality}
    \vspace{-5mm}
\end{figure*}

In assessing the reality gap, it is important to note that while alternative comparison methods exist, we selected this approach for its robustness. As depicted in Fig.\ref{fig:odometry_chart}, we modeled angular velocity as a proposed function, complemented by a constant linear speed. The robot's trajectory, whether following a straight path or engaging in oscillatory motion, was determined by these parameters. The resulting path was a pentagon, chosen for its geometric complexity to challenge the robot's navigation systems.

The simulated robot's trajectory demonstrated a high degree of congruence with the real robot's odometry, showcasing the simulator's precision. The slight discrepancies observed are primarily due to modeling simplifications which are part of the current simulation design. These differences provide valuable insights, guiding the refinement of our simulation parameters to more intricately mirror real-world dynamics while maintaining computational efficiency.


\subsection{Reasoning Process and Human-Robot Interaction}

The integration of LLMs into YamaS signifies a leap forward in AI-driven robotic behavior generation. The LLM agent receives a rich, structured description of the environment, an intricate input that informs its decision-making process. This procedure is exemplified in Fig.~\ref{fig:virtual_reality}, where the agent is tasked with interpreting complex environment data to strategize and perform actions such as object retrieval and placement.

The following is a streamlined representation of the LLM agent's actions and reasoning process:

\begin{minipage}{0.95\linewidth}
\vspace{2mm}
\color{darkgray}
    \textbf{\underline{Environment Description - LLM Inputs}}
    
    \vspace{2mm}

    \textbf{[Input]}: Received areas information: Area 1 has 1 Orange Ball, 1 Red Zone, 1 Green Zone, 5 obstacles.

    \textbf{\underline{Activity Information  - LLM Outputs}}

    \vspace{2mm}
    
    \textbf{LLM Agent [planning]}: Reasoning:
    \begin{enumerate}
        \item search\_ball(``Orange") to find the Orange Ball.
        \item catch\_the\_ball(``Orange") to pick up the Orange Ball.
        \item search\_zone(``Green") to find the Green Zone.
        \item go\_to\_zone(``Green") to move towards the Green Zone.
        \item leave\_ball() to leave the Orange Ball in the Green Zone.
    \end{enumerate}
    
    \textbf{LLM Agent [answering]}: Response: ``I will search for and catch the Orange Ball, then find and go to the Green Zone to leave the ball there."
    
    \textbf{LLM Agent [calling\_function]}: Tasks to be executed: search\_ball(`Orange'); catch\_the\_ball(`Orange'); search\_zone(`Green'); go\_to\_zone(`Green'); leave\_ball();
\end{minipage}

Analytically, the LLM's inputs consist of detailed environmental elements. Including object types, locations, and obstacles, formatted to facilitate the model's comprehension and planning. The agent's outputs, which are its planned actions and reasoning, are direct responses to these inputs. They reflect the agent's capacity to parse and understand its surroundings, select appropriate actions, and sequence them to achieve a goal efficiently.

These results underline the potential of YamaS as a research tool in robotics, presenting a realistic and sophisticated platform where both robots and algorithms can be tested and refined before being deployed in the real world.

%% file: sessions/5_conclusions.tex
\section{Conclusion}\label{section:conclusions}


Reflecting on the functionalities and experiments detailed in this paper, the YamaS simulation framework represents a significant step forward in the domain of robotics simulation. While it has been instrumental in refining the simulated behavior of the Yamabiko robot to closely match its real-world counterpart, we recognize that there is always room for improvement. The integration of LLMs and procedural generation technologies has streamlined the creation and testing of robotic behaviors in diverse scenarios, and the addition of VR has deepened the potential for immersive HRI studies.

Crucially, the YamaS platform serves as a bridge connecting theoretical robotics research with practical, real-world application, and it does so with a measured understanding of its current scope and limitations. Future work appears promising, especially with the potential of applying LLMs more directly to robot control systems, inspired by pioneering works such as the Voyager algorithm~\cite{wang2023voyager}. This direction holds the potential to accelerate the robot simulations' step towards more autonomous and intelligent robotic systems. 